\documentclass[sigconf]{acmart}

\usepackage{bbm}
\usepackage{graphicx}
\usepackage{amsthm}
\usepackage{pifont}
\usepackage{diagbox}
\usepackage{amsmath}

\usepackage{amssymb}
\usepackage{amsfonts}
\usepackage{booktabs}
\usepackage{multirow}
\usepackage{subfigure}

\AtBeginDocument{%
  \providecommand\BibTeX{{%
    \normalfont B\kern-0.5em{\scshape i\kern-0.25em b}\kern-0.8em\TeX}}}

\setcopyright{acmcopyright}
\copyrightyear{2022}
\acmYear{2022}
\acmDOI{10.1145/1122445.1122456}

\acmConference[MM`22]{Proceedings of the 30th ACM International Conference on Multimedia}{October 10--14, 2022}{Lisbon, Portugal}
\acmPrice{15.00}
\acmISBN{978-1-4503-XXXX-X/18/06}

\begin{document}

%%
%% The "title" command has an optional parameter,
%% allowing the author to define a "short title" to be used in page headers. Peers
\title{Layout-Aware Information Extraction for Document-Grounded Dialogue: Dataset, Method and Demonstration}
\renewcommand{\shorttitle}{Layout-Aware Information Extraction for Document-Grounded Dialogue}
%%
%% The "author" command and its associated commands are used to define
%% the authors and their affiliations.
%% Of note is the shared affiliation of the first two authors, and the
%% "authornote" and "authornotemark" commands
%% used to denote shared contribution to the research.
%\author{Anonymous Authors}
%\email{trovato@corporation.com}
%\orcid{1234-5678-9012}
%\author{Zhenyu Zhang$^{12\dag}$, Bowen Yu$^{123\dag}$, Haiyang Yu$^{3}$, Tingwen Liu$^{12\ddag}$,}
%\author{Cheng Fu$^{3}$, Jingyang Li$^{3}$, Chengguang Tang$^{3}$, Jian Sun$^{3}$}
%\thanks{$^\dag$ Both authors contributed equally to this research.}
%\thanks{$^\ddag$ Corresponding author.}
%\affiliation{
%  \institution{$^{1}$Institute of Information Engineering, Chinese Academy of Sciences}
%  \institution{$^{2}$School of Cyber Security, University of Chinese Academy of Sciences}
%  \institution{$^{3}$DAMO Academy, Alibaba Group}
%}
%\email{{zhangzhenyu1996, yubowen, liutingwen}@iie.ac.cn}
%\email{{yifei.yhy, fucheng.fuc, qiwei.ljy, chengguang.tcg, jian.sun}@alibaba-inc.com}

\author{Zhenyu Zhang}
\authornote{Both authors contributed equally to this research.}
\affiliation{%
  \institution{Institute of Information Engineering, Chinese Academy of Sciences \\ School of Cyber Security, University of Chinese Academy of Sciences}
%   \city{Beijing}
%   \country{China}
  }
\email{zhangzhenyu1996@iie.ac.cn}

\author{Bowen Yu}
\authornotemark[1]
\affiliation{
  \institution{DAMO Academy, \\ Alibaba Group\\
  Institute of Information Engineering, Chinese Academy of Sciences}
%   \city{Beijing}
%   \country{China}
  }
\email{yubowen.ybw@alibaba-inc.com}

\author{Haiyang Yu}
\affiliation{%
  \institution{DAMO Academy, \\ Alibaba Group}
%   \city{Beijing}
%   \country{China}
  }
\email{yifei.yhy@alibaba-inc.com}

\author{Tingwen Liu}
\authornote{Corresponding author.}
\affiliation{%
  \institution{Institute of Information Engineering, Chinese Academy of Sciences \\ School of Cyber Security, University of Chinese Academy of Sciences}
%   \city{Beijing}
%   \country{China}
  }
\email{liutingwen@iie.ac.cn}

\author{Cheng Fu, Jingyang Li}
\affiliation{%
  \institution{DAMO Academy, \\ Alibaba Group}
%   \city{Beijing}
%   \country{China}
  }
\email{fucheng.fuc@alibaba-inc.com}
\email{qiwei.ljy@alibaba-inc.com}

\author{Chengguang Tang, Jian Sun, Yongbin Li}
\affiliation{%
  \institution{DAMO Academy, \\ Alibaba Group}
%   \city{Beijing}
%   \country{China}
  }
\email{chengguang.tcg@alibaba-inc.com}
\email{jian.sun@alibaba-inc.com}
\email{shuide.lyb@alibaba-inc.com}
%%
%% By default, the full list of authors will be used in the page
%% headers. Often, this list is too long, and will overlap
%% other information printed in the page headers. This command allows
%% the author to define a more concise list
%% of authors' names for this purpose.
\renewcommand{\shortauthors}{Zhang et al.}

%%
%% The abstract is a short summary of the work to be presented in the
%% article.
\begin{abstract}

Building document-grounded dialogue systems have received growing interest as documents convey a wealth of human knowledge and commonly exist in enterprises.
Wherein, how to comprehend and retrieve information from documents is a challenging research problem.
Previous work ignores the visual property of documents and treats them as plain text, resulting in incomplete modality.
In this paper, we propose a Layout-aware document-level Information Extraction dataset, LIE, to facilitate the study of extracting both structural and semantic knowledge from visually rich documents (VRDs), so as to generate accurate responses in dialogue systems.
LIE contains 62k annotations of three extraction tasks from 4,061 pages in product and official documents, becoming the largest VRD-based information extraction dataset to the best of our knowledge.
We also develop benchmark methods that extend the token-based language model to consider layout features like humans. 
Empirical results show that layout is critical for VRD-based extraction, and system demonstration also verifies that the extracted knowledge can help locate the answers that users care about.
% The dataset and baselines are publicly available at https://github.com/xxx/xxx.
\end{abstract}

%In recent years, the methods based on pre training language model and DOM tree have made great progress in the extraction task, but their eyes are always limited to a single web page and do not see the cross web page co-occurrence between elements.
%We find that in the DOM tree corresponding to different web pages, the elements in the same position usually play the same semantic role and have similar characters and expressions. In this paper, we propose a new method to effectively use structured co-occurrence

%%
%% The code below is generated by the tool at http://dl.acm.org/ccs.cfm.
%% Please copy and paste the code instead of the example below.
%%


%% Keywords. The author(s) should pick words that accurately describe
%% the work being presented. Separate the keywords with commas.
%%
%% This command processes the author and affiliation and title
%% information and builds the first part of the formatted document.
\maketitle

\section{Introduction}

Dialogue systems are playing an increasingly important role in various business applications~\cite{wei2018task,he2022galaxy,he2022sapce}.
Currently, enterprises and organizations often own abundant business documents, which have the potential to solve user queries, such as product documentation and policy guidance. 
However, users still prefer to efficiently obtain the desired answer through interactive dialogue, rather than inefficient and inconvenient retrieval in lengthy documents.
Therefore, building machine-assisted agents based on massive documents at low cost is a topic of great concern in industry~\cite{choi2018quac,campos2020doqa,chen2020bridging,zhang2021aware}.

Towards this goal, the document-grounded dialogue (doc2dial) system is usually decomposed into two sub-tasks: knowledge extraction and response generation~\cite{feng2020doc2dialframe}.
Specifically, knowledge extraction aims to identify the most relevant knowledge in the associated document, and response generation focuses on generating a fluent response.
Wherein, the first step is the key point that drives doc2dial different from other dialogue tasks.
However, in this part, previous works typically assume the input to be textual strings~\cite{chen2021building}, while many real-world business documents are scanned or digital-born (e.g., images of invoices, forms in PDF format).
\emph{When converting these visually rich documents (VRDs) into snippets of plain text, a lot of layout information is discarded, which is exactly the indicative signal to help humans quickly understand the document and search for some interested knowledge.}

Based on an inductive analysis of a large number of real-world queries in business doc2dial systems, we conclude that the answers that users care about, as shown in Figure 1, can be roughly summarized into two kinds of knowledge in the document: (1) \emph{coarse-grained description knowledge} at section level, (2) \emph{fine-grained fact knowledge} within a section.
Besides, to accurately locate the section related to the query for better finding relevant knowledge, the system also needs to be aware of the \emph{hierarchy knowledge} of a document.
Recently, utilizing layout information to help extract information from VRDs has attracted significant research interest, and many layout-aware datasets are proposed.
For example,
CORD~\cite{park2019cord} and SROIE~\cite{huang2019icdar2019} aim to extract values for pre-defined keys from receipts, 
FUNSD~\cite{jaume2019funsd} and EPHOIE~\cite{wang2021towards} focus on the automatic extraction of key-value pairs from form-like documents.
\emph{However, the task forms defined in these datasets are within the scope of span extraction and ignore the extraction of fact knowledge.}

To overcome the above limitations, we construct a new Layout-aware dataset for document-level Information Extraction, LIE.
Unlike previous text-only or form-like datasets, LIE is built on multi-page VRDs, requiring systems not only to extract specific text spans but also the relations among them.
We provide the original file and all tokens with spatial positions (i.e., layout feature) for each document, then release three information extraction (IE) tasks: \emph{hierarchy extraction}, \emph{section extraction} and \emph{relation extraction}, where the first one is to structuralize the multi-page document, and the last two aims to collect specific knowledge described above.

We also propose benchmark methods for the new dataset.
Inspired by the mechanism that humans understand VRDs through both text semantics and document layout, we extend popular token-based language models with layout features.
Similar to the 1D position embeddings in transformer-based models, we introduce 2D layout embeddings to capture the spatial relationship among tokens within a document. 
Then the layout-aware model is firstly pre-trained on unlabeled data with two self-supervised tasks, masked layout-language model and potential heading selection, to better learn the newly introduced layout embeddings, then fine-tuned on task-specific labeled data.
Experimental results suggest that injecting layout could improve the IE performance significantly, and the pre-training process accelerates the convergence of models on downstream tasks.
Moreover, our model has already gone into production in a world-leading cloud computing platform, we also demonstrate the workflow of doc2dial service and how layout-aware extraction plays a vital role at the end of the paper.

\begin{figure}[t]
\centering
\includegraphics[width=\linewidth]{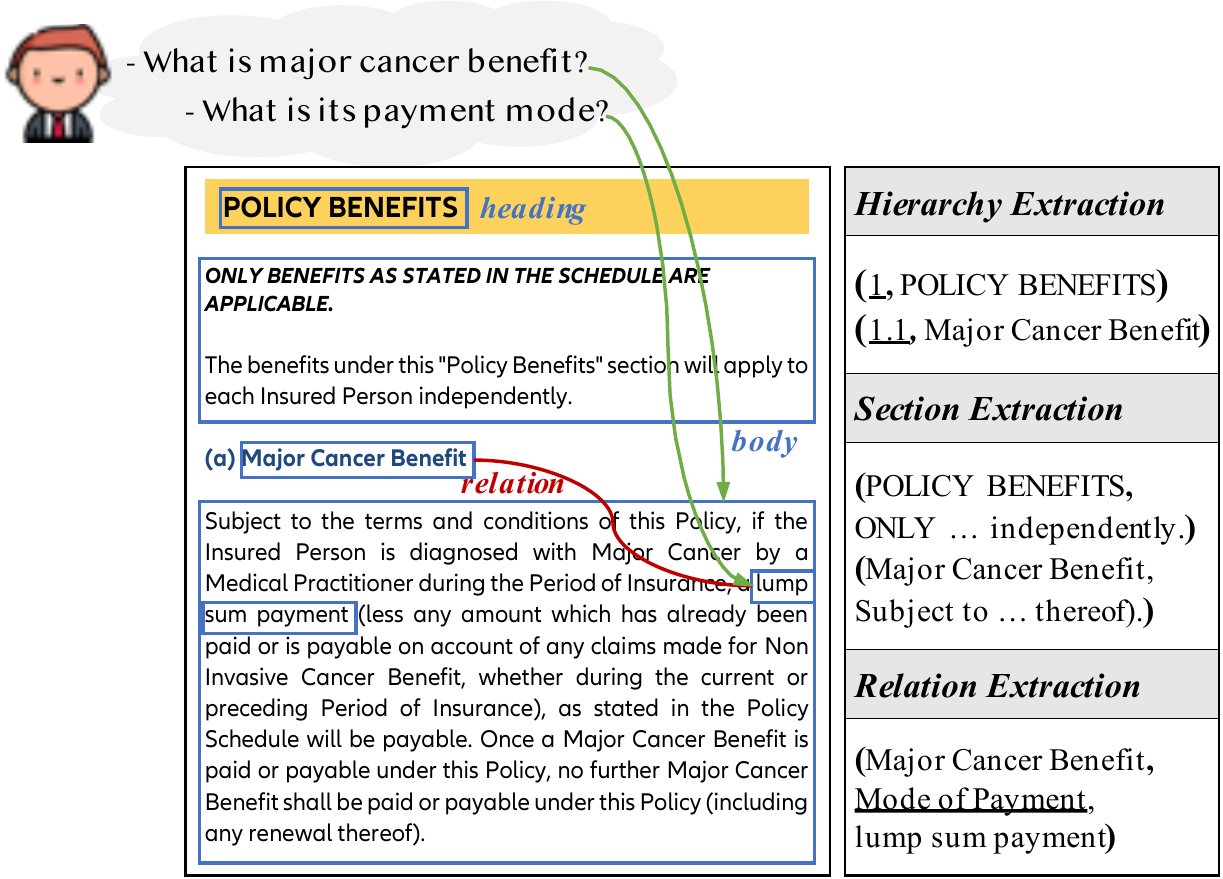}
\caption{The illustration of doc2dial, VRD, and the proposed IE tasks (from left to right). Best viewed in color.}
\label{fig.example}
\end{figure}

\section{Related Work}

\noindent\textbf{Document-based Information Extraction.}
Most work in early phase IE focuses on the sentence-level.
Nowadays, many researchers extend the scope of extraction beyond isolated sentences~\cite{hendrickx2010semeval,zhang2020document,zhou2022survey}.
\citet{yao2019docred} introduced a cross-sentence relation extraction dataset, DocRED, based on Wikipedia.
\citet{yu2020dialogue} proposed DialogRE to support the prediction of relations in multi-turn dialogues.
The length of these documents is equivalent to DocRED, which is actually the paragraph-level extraction.
SciREX~\cite{jain2020scirex} releases the task of identifying the main results of a scientific article.
However, the source document provides more layout information than a sentence or paragraph. Most existing IE datasets drop these valuable features in the process of text reading.
Recently, many datasets for IE from VRDs are proposed.
CORD~\cite{park2019cord} proposes the task to label each word in receipts to the right field, wherein $30$ fields under $4$ categories are defined.
SROIE~\cite{huang2019icdar2019} aims to extract values from each receipt for $4$ pre-defined keys.
Towards extracting keys from documents,
FUNSD~\cite{jaume2019funsd} is the most widely-used dataset and constructed on various business forms with $4$ keys.
EPHOIE~\cite{wang2021towards} is constructed based on the images of examination paper heads, and $5$ kinds of key-value pairs need to be extracted.
What is most different about our dataset is that these only support span extraction and their documents are form-like with few words. 
Our work focuses more on \emph{enabling machines to handle the same multimodal input as humans when they read real-world documents} and \emph{extracting various structural and semantic knowledge from VRDs so as to drive better doc2dial systems}.

\noindent\textbf{Layout-aware Language Model.} 
Humans perceive the document through many aspects, such as language, layout, and vision. 
Based on the powerful modeling ability of Transformer, LayoutLM~\cite{xu2020layoutlm} and LayoutLMv2~\cite{xu2021layoutlmv2} are successively proposed to learn the multi-modality interaction and achieve great success.
Following the paradigm, 
StructuralLM~\cite{li2021structurallm} leveraged cells and layout to make the model aware of which words are from the same cell.
StrucTexT~\cite{li2021structext} extracted semantic features from different levels and modalities.
DocFormer~\cite{appalaraju2021docformer} incorporated multi-modality self-attention with shared spatial embeddings.
Besides,
\citet{wei2020robust} combined the power of language models and graph neural networks for the extraction from VRDs.
\citet{zhang2020trie} presented an end-to-end network to bridge text reading and IE for document understanding.
The motivation behind these studies is similar to ours, but they pay too much attention to layout and ignore the in-depth semantics. 
Our work aims to \emph{go beyond surface modeling and understand the document logical hierarchy from a semantic perspective}. 

\noindent\textbf{Document-grounded Dialogue.}
The task of reading documents and responding to queries has been the trigger of recent research advances~\cite{choi2018quac,wei2018task,campos2020doqa}.
Recently, the doc2dial dataset~\cite{feng2020doc2dialdata} is proposed to facilitate the frontier exploration and landing application of document-grounded dialogue systems. It contains 4,793 goal-oriented dialogues and a total of 488 associated grounding documents from 4 domains for social welfare.
The proof-of-concept doc2dial framework~\cite{feng2020doc2dialframe} and subsequent attempts~\cite{fadnis2021doc2bot,chen2021building,khosla2021team} are following the knowledge extraction - response generation paradigm.
However, their data pre-processing includes a well-designed text reading step, which draws the government document into the form of plain text.
In this paper, we pay more attention to the step of knowledge extraction and argue that \emph{the multimodal features of business documents should not be lightly dismissed}.

\begin{figure*}[t]
\centering
\subfigure[document length (pages).]{
\includegraphics[width=0.23\linewidth]{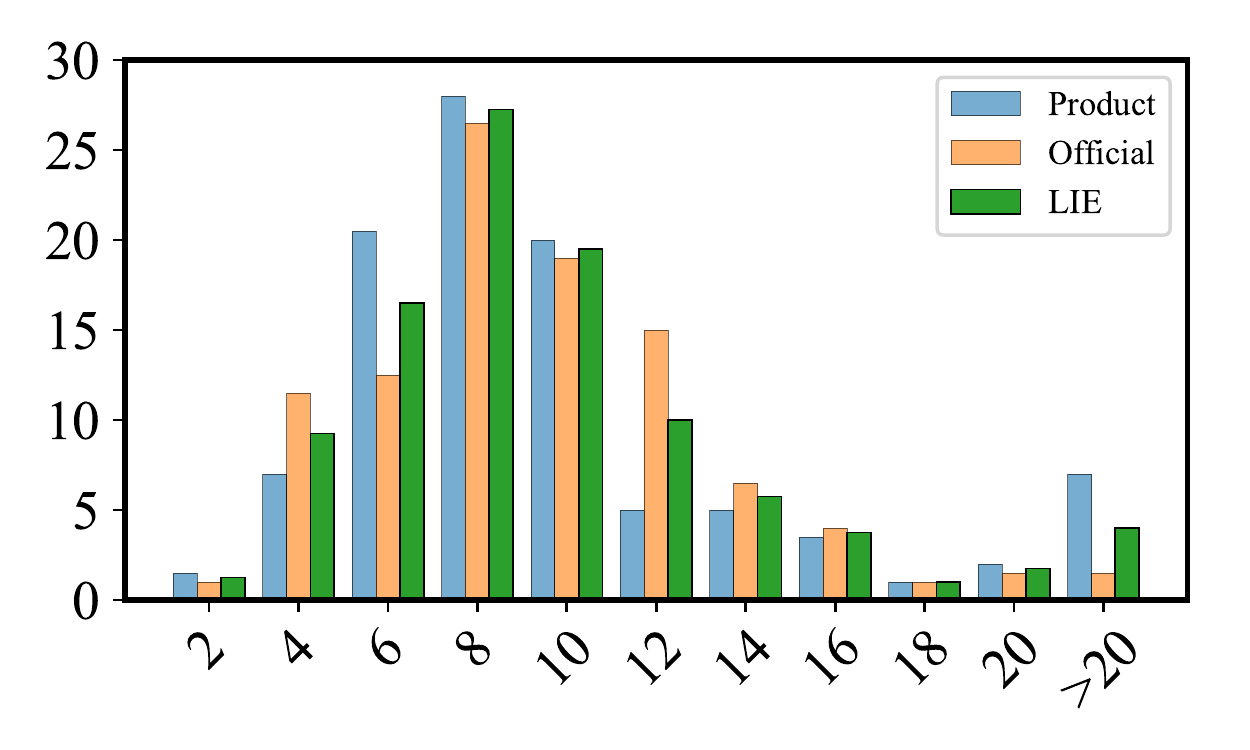}
\label{fig.dist.page}
}
\subfigure[document length (words).]{
\includegraphics[width=0.23\linewidth]{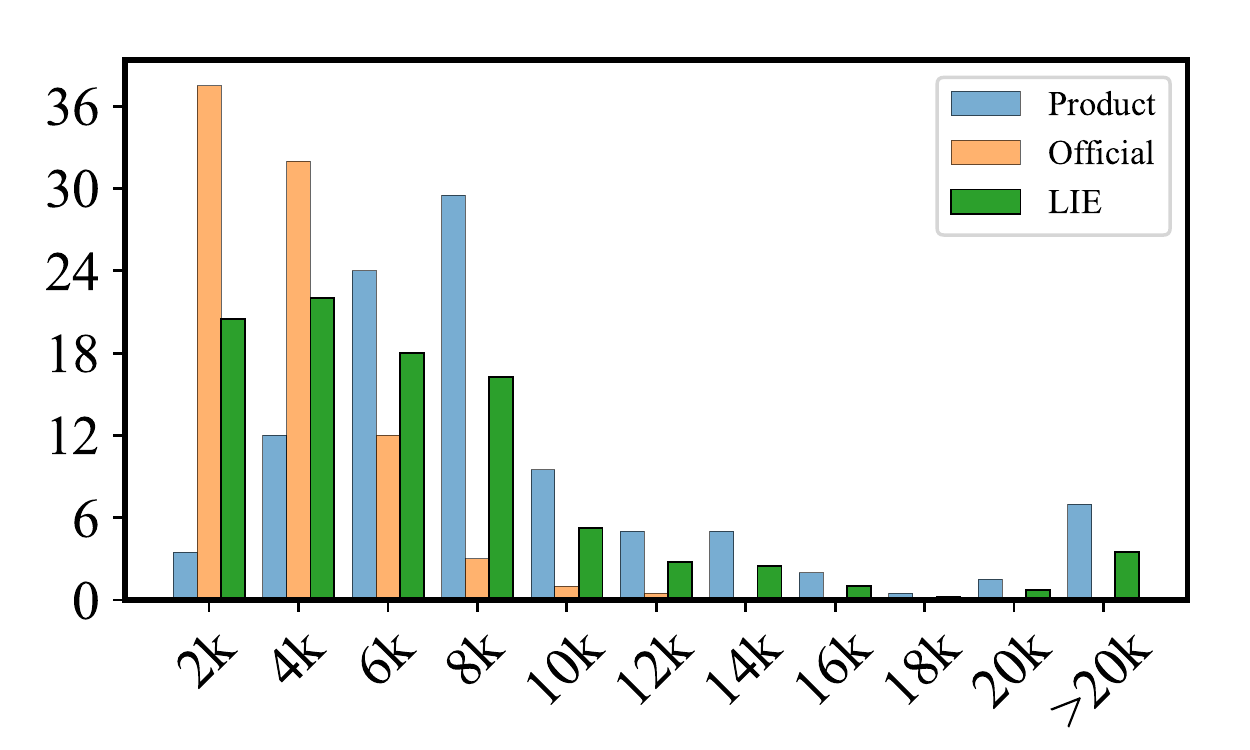}
\label{fig.dist.word}
}
\subfigure[section length (words).]{
\includegraphics[width=0.23\linewidth]{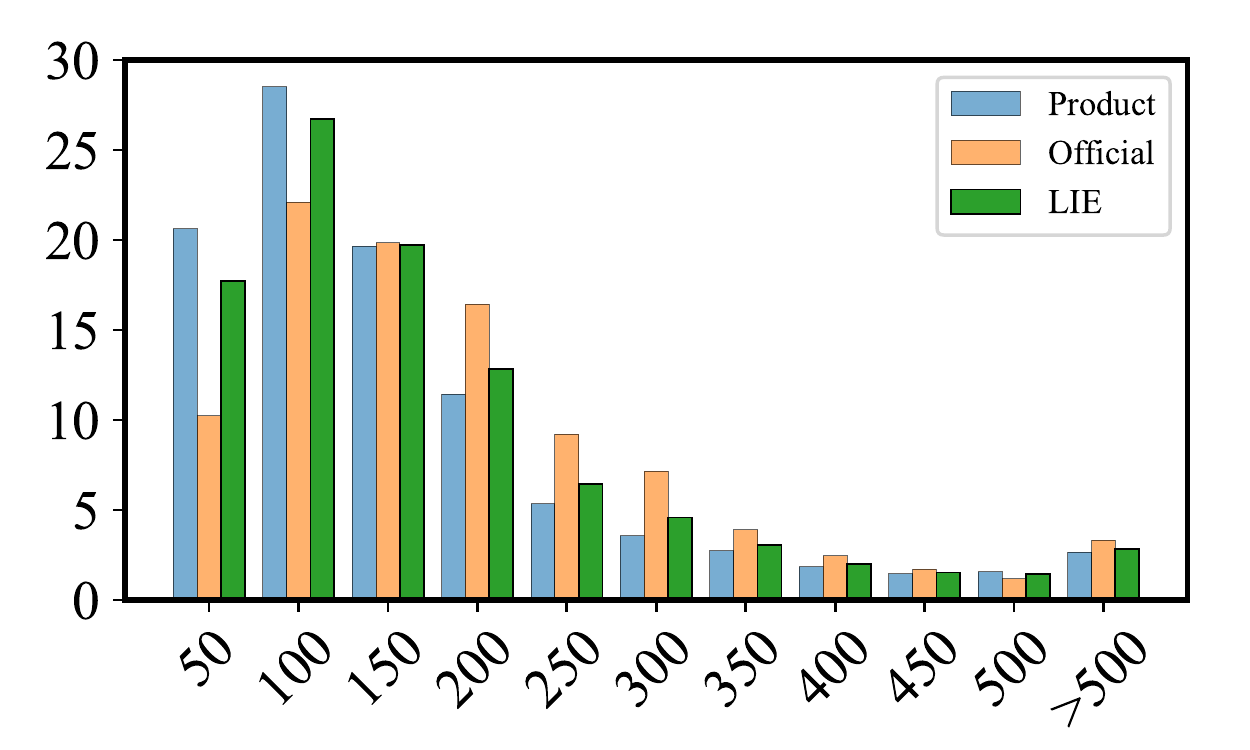}
\label{fig.dist.section}
}
\subfigure[argument distance (words).]{
\includegraphics[width=0.23\linewidth]{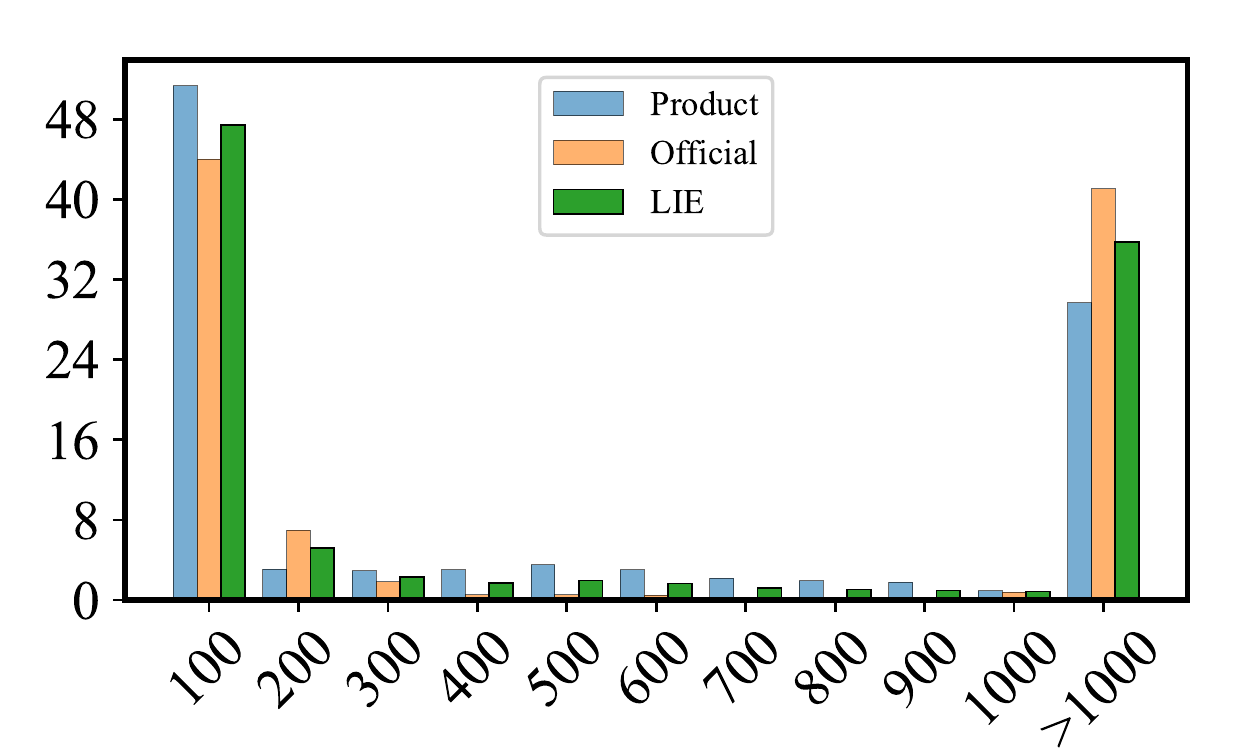}
\label{fig.dist.args}
}
\caption{Distribution of document length, section length, and argument distance in LIE, where the y-axis is in percent. }
\label{fig.dist}
\end{figure*}

\section{Dataset}

With the development of learning-based algorithms, a comprehensive benchmark conducted for specific tasks is a prerequisite to motivate advanced works.
 Compared with previous IE datasets, LIE is formed with several desirable properties: 
 (i) It is proposed for the extraction from multi-page VRDs, which brings new opportunities and challenges than previous text-only or form-like settings. 
 (ii) It pays more attention to the extraction of relational fact knowledge, while prior work only focuses on extracting salient spans.
 (iii) It needs to understand and parse the hierarchical structure of documents, which is able to facilitate kinds of document understanding tasks and follow-up document-grounded applications.

\subsection{Task Formulation}

To better extract valuable knowledge from VRDs and support the downstream doc2dial systems, we formulate three IE tasks:

\noindent\textbf{Hierarchy Extraction (HE).} 
It aims to extract document logical hierarchy and generate table-of-content (TOC) by organizing section headings in order.
An implementable goal is to detect \emph{section headings} with their \emph{global levels}.
In Figure~\ref{fig.example}, ``\verb|POLICY| \verb|BENEFITS|'' is the first section heading with level $1$, so the TOC number is $1$, ``\verb|Major| \verb|Cancer| \verb|Benefit|'' is a subsection heading with level $2$ and number $1.1$.
Typically, headings at the same level share the same font formats, meaning that the non-text signals, including type, size, and indentation, are vital for the task. 

\noindent\textbf{Section Extraction (SE).} 
It could be regarded as an extension of key-values extraction in previous form-like datasets~\cite{jaume2019funsd}, where the heading of a section is key, and the corresponding body is value.
For VRDs, bodies are usually below or to the right of headings and sometimes indented. 
Figure~\ref{fig.example} shows concrete examples.
When mounting the (\emph{heading}, \emph{body}) pairs on TOC, we can easily transform a document into its hierarchical tree, which is beneficial to many downstream applications~\cite{wan2021does}.

\noindent\textbf{Relation Extraction (RE).} 
It needs to detect \emph{two entities} and identify their \emph{relations} from VRDs~\cite{yu2020joint, wang2020tplinker}, as shown in Figure~\ref{fig.example}. 
When converting VRDs into plain text, two entities may become far away, but actually, they may be very close or even aligned in the document.
In this task, relations belong to a pre-defined set, so it is suitable for obtaining structured knowledge that we care about.

\begin{table}[t]\small
	\centering
	\caption{Layout-aware IE datasets comparison. \#Doc. Length refers to the average page numbers.}
	\begin{tabular}{lcccccc}
		\toprule
		  Datasets & \#Pages  & \#Doc. Length & Fact Extraction? \\
		\midrule
		CORD~(\citeyear{park2019cord})  & 1000 & 1  & \ding{55} \\
		SROIE~(\citeyear{huang2019icdar2019}) & 973 & 1  & \ding{55}\\
		FUNSD~(\citeyear{jaume2019funsd}) & 199 & 1   & \ding{55}\\
		EPHOIE~(\citeyear{wang2021towards}) & 1494 & 1  & \ding{55}\\
		\midrule
		LIE (our) & \textbf{4061} & \textbf{10.15} & \ding{51} \\
		\bottomrule
	\end{tabular}
	\label{tab.comp}
\end{table}

\subsection{Data Collection and Annotation}

To construct LIE, we choose two representative VRDs in the real world, i.e., \emph{product documents} and \emph{official documents}.
Specifically, we download Chinese product documents in PDF format from portal websites of the insurance sector and collect official documents issued by government departments from web search engines. 
Different from previous VRD-based datasets, these documents contain more textual fragments, which is more conducive to evaluating the language understanding ability. 
Furthermore, the semantic information is expressed not only through the text in each fragment but also how the fragments organized, so it is still important to perceive the document layout. 

Next, we utilize PDFPlumber\footnote{\url{https://github.com/jsvine/pdfplumber}} to extract tokens with bounding boxes automatically.
Here, a document page is considered as a 2D coordinate system with the \emph{top-left origin}.
Therefore, in the parsing result, each word is aligned with a unique quadruplet ($x_0, y_0$, $x_1, y_1$), where ($x_0, y_0$) corresponds to the position of the upper left in bounding box, and ($x_1, y_1$) represents the lower right.
We also provide the PDF file so that downstream models have the opportunity to access rich original information, such as visual features.

The data annotation process is carried out by crowdsourcing.
Annotators are provided with PDF files and asked to fill the slots by copying text spans (e.g., headings) or generating (i.e., TOC) numbers according to the annotation specification.
From the practical application perspective, we summarize 18 and 15 relations for product documents and official documents, respectively, as  pre-defined relation schemes.
%(see also Table~\ref{tab.schema} in Appendix)
All the (20) annotators have linguistic knowledge, are instructed with formal annotation principles, and pass trial annotation.
To ensure the labeling quality, each instance is labeled by at least two annotators. If the two annotators have disagreements on an instance, it will be assigned to a third annotator.
The annotation results are also randomly checked. If the accuracy is lower than 95\% (measured in the document-level), all results of the annotator will be reviewed.
We ensure that all the annotators are fairly compensated by market price according to their workload.
Finally, we develop heuristic rules to align the parsing results with annotations, and about 2\% of the annotation instances cannot be aligned. We discard them and blame them for the parsing error.
% (see also Appendix for more details).

\subsection{Data Statistics and Analysis}

Using the annotation procedure mentioned above, we build a dataset of 4,061 fully annotated pages from 400 documents (200 product and official documents, respectively). All documents have accompanied layout features.
Note that these documents totally come from more than 150 organizations, and documents from one organization usually have more than one layout format, \emph{making the dataset diverse enough and robust to practical applications}.

Table~\ref{tab.comp} provides the statistics of LIE and some representative VRDs-based IE datasets.
One can observe that LIE is significantly larger than all these datasets and is the only dataset that supports extracting triplet knowledge from consecutive pages. 
In contrast, previous datasets are built based on single-page documents and focus to extract unary spans.
We hope the large-scale LIE dataset with complete input modality and task formulation could drive VRDs-based IE from form understanding forward to accurate document understanding.
We analyze LIE in detail and plot the data distribution in Figure~\ref{fig.dist} to support the follow-up study.
Figure~\ref{fig.dist.page} and~\ref{fig.dist.word} depict the distribution of pages and words in a document, respectively, and Figure~\ref{fig.dist.section} show the distribution of section length.
Obviously, the documents have a large number of pages and words, and even one section may exceed 500 words.
The distance distribution between two entity arguments in relation extraction is drawn in Figure~\ref{fig.dist.args}, an intuitive observation is that about half of the arguments are far away in text sequence, or even not in the same section.
By and large, all the above are beyond the focus of most existing IE benchmarks and pose new challenges to document understanding and extraction modules.

\begin{figure}[t]
\centering
\includegraphics[width=\linewidth]{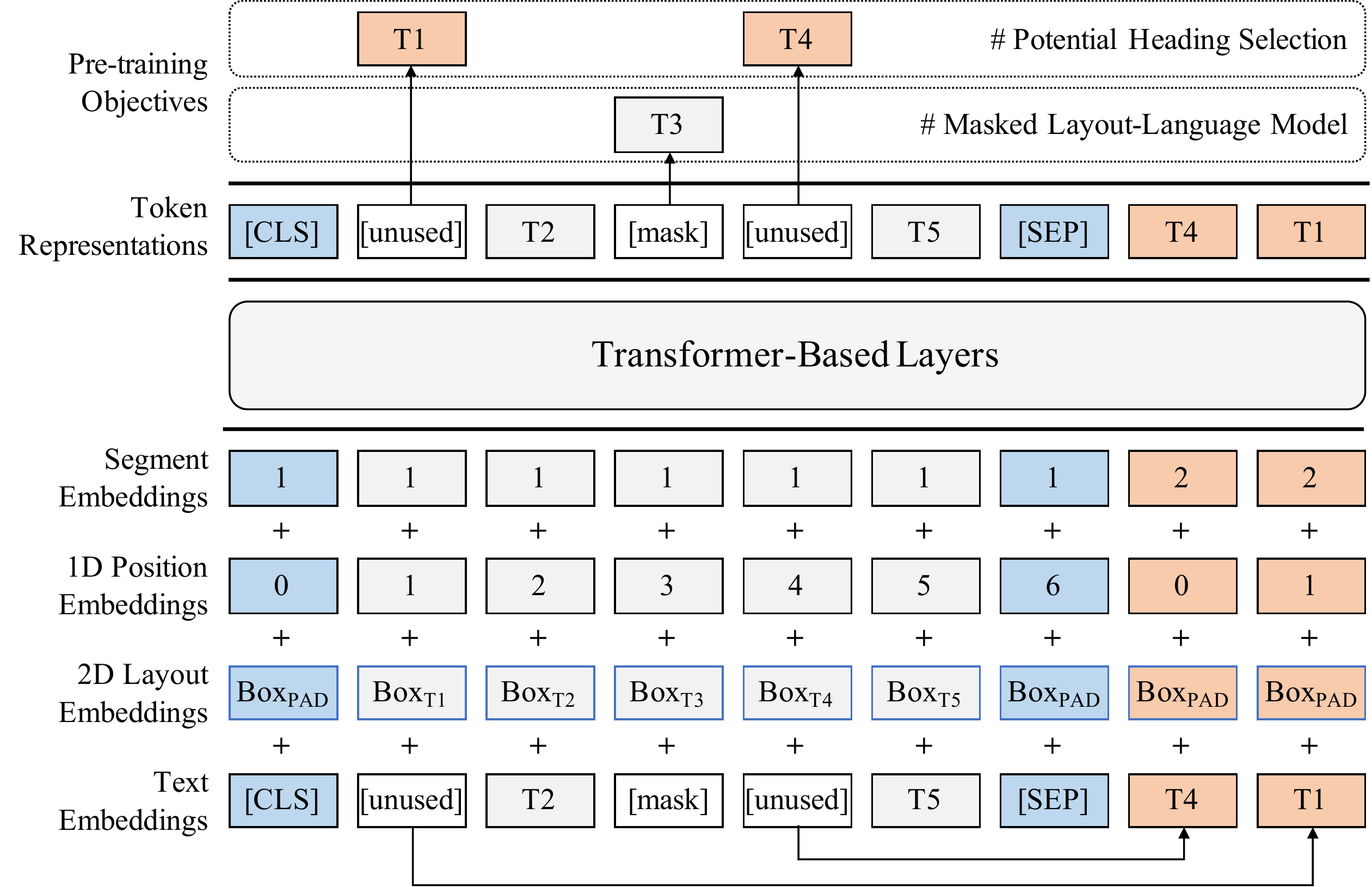}
\caption{Diagram of the layout-aware language model and pre-training strategies, where 2D layout embeddings are integrated into input layer and potential head selection is proposed to learn document structure. T1 and T4 refer to the selected potential headings in the pre-training stage, we simplify them into one token for a good visualization.}
\label{fig.model}
\end{figure}

\section{Method}

Different from plain text, there is rich semantic knowledge hidden under the textual format and layout structure in VRDs, so traditional token-based language models are not directly applicable. 
In this section, we incorporate 2D layout features into transformer-based models and further propose specific solutions for the IE tasks.

\subsection{Base Architecture}

Figure~\ref{fig.model} shows the architecture of our document encoder.
Similar to LayoutLMv2~\cite{xu2021layoutlmv2}, we built the document encoder in a layout-aware manner, which accepts information from both text and layout.
Beyond the 1D position embedding that models the sequential knowledge, we introduce 2D layout embedding to capture the spatial position in documents. 
We normalize and discretize all 2D coordinates to integers in the range [0,1000] and lookup \emph{four} layout embeddings from \emph{two} embedding tables, to embed $x$-axis and $y$-axis features separately. 
In the end, the textual embedding is added with segment, position, and layout embeddings to get the ultimate input of the transformer encoder. 
Based on the self-attention mechanism, embedding 2D knowledge into language models will better align layout features with the semantic representation.
Finally, the output contextual representation of the transformer can be utilized for the following task-specific layers.

\subsection{Sequence Labeling Solution}

To quantitatively evaluate the challenges of LIE, 
we propose a set of solutions for specific tasks whose design philosophy is trying to ensure simplicity.
Here, we unify all the tasks into a sequence labeling framework with different tagging schemes.

\noindent\textbf{Hierarchy Extraction (HE).} 
In LIE, there are usually multiple articles within a document. 
Here, we divide the headings in one document into four levels: article headlines (\verb|L0|) and section headings at different hierarchies  (\verb|L1/2/3|).
The tag set is defined as \{\verb|B-H|, \verb|E-H|, \verb|O|\}, in which \verb|B| and \verb|E| mark the beginning and end of a heading, \verb|H| $\in$ \{\verb|L0,L1,L2,L3|\} encodes its level.
\verb|O| represents the other tag, indicating that the word is independent of the extracted results.
To determine the TOC of documents, we decode each heading span by way of boundary nearest matching~\cite{zheng2017joint} and number them in order according to their levels and positions.

\noindent\textbf{Section Extraction (SE).} 
Similarly, we label the spans of section headings and bodies with the tag set \{\verb|B-H|, \verb|B-B|, \verb|E-H|, \verb|E-B|, \verb|O|\}.
In the decoding process, headings and bodies are combined into a section based on the nearest principle.

\noindent\textbf{Relation Extraction (RE).} 
The labeling model is quite more complex than that in section extraction as it needs to extract two entities, match them, and determine their relations.
We design a tripart tagging scheme \{\verb|B-S|, \verb|E-S|, \verb|B-O-Rel|, \verb|E-O-Rel|, \verb|O|\}, where \verb|S| and \verb|O| denote the subject and object entities, \verb|Rel|$\in \mathcal{R}$ denotes the relation from pre-defined set.
Thus, the total number of tags is $2\times |\mathcal{R}|+3$.
During decoding, we first detect the subject entity, then the object entity with its involved relation, and finally pair them according to the nearest principle.

\subsection{Pre-Training Strategy}

Inspired by the great success of pre-trained language models, we pre-train the base encoder, to optimize the newly introduced parameters (i.e., layout embedding tables) and transfer token-based models to the visually rich dataset, with the following self-supervised objectives. 
Here, 100$k$ additional pages from 10$k$ in-domain documents (5$k$ product and official documents, respectively) are collected and parsed as the pre-training corpus.

\noindent\textbf{Masked Layout-Language Model (MLLM).}
Similar to the famous masked language model task in BERT~\cite{devlin2019bert}, the objective of MLLM is to recover the masked text token based on its text context and whole layout clues.
The layout embeddings remain unchanged for masked token, which means that the model knows each masked token’s location on the page, thereby equipping the token-based model with layout awareness.

\noindent\textbf{Potential Heading Selection (PHS).}
To help the model learn structural and semantic knowledge among text fragments, we propose the PHS task.
Specifically, we randomly select 15\% of the training instances and develop heuristic rules to find text fragments with significant visual features (e.g., large font size) as potential headings.
Next, these fragments are masked in the original sequence and spliced to the end of the sequence after random scrambling.
Similarly, only text embeddings are masked and layout embeddings are retained.
During pre-training, the goal is to select the most appropriate fragment for each masked position. 
We calculate the cross-entropy loss in optimization.
The key intuition behind this is that \emph{an eye-catching text fragment is usually a potential heading}, which semantically dominates the body of a section.
In this way, the model could learn better with multi-modality clues and prompt the understanding of long documents.

\section{Experiments}

To maximize the reusability of LIE, we provide an official partition to split annotated documents into train, dev, and test sets in the ratio of 6:2:2, and report the statistics in Table~\ref{tab.stat}.
In this section, we conduct comprehensive experiments to evaluate the new benchmark, and also discuss some possible future directions for VRDs-based IE.

\begin{table}[t]\small
	\centering
	\caption{Statistics of the LIE train/dev/test datasets.}
	\begin{tabular}{lccccccccc}
		\toprule
		  & & Hierarchy & Section & Relation \\
		  & & Extraction & Extraction & Extraction \\
		\midrule
		\multirow{2}{*}{Train} & \emph{Product}  & 7,590 & 6,561 & 2,034 \\
		                       & \emph{Official} & 2,830 & 2,390 & 2,081 \\
		\midrule
        \multirow{2}{*}{Dev}   & \emph{Product}  & 2,482 & 2,183 & 655 \\
		                       & \emph{Official} & 1,068 & 892 & 771 \\
		\midrule
		\multirow{2}{*}{Test}  & \emph{Product}  & 2,368 & 2,065 & 644 \\
		                       & \emph{Official} & 1,031 & 861 & 717 \\
		\bottomrule
	\end{tabular}
	\label{tab.stat}
\end{table}

\subsection{Experiment Setup}

\textbf{Implementation Details.}
We develop our model based on Transformers~\cite{wolf2020transformers} and employ BERT~\cite{devlin2019bert} as backbone, with the official \emph{bert-base-chinese} model.
%\footnote{\url{https://huggingface.co/bert-base-chinese}}
The input sequence is trimmed to a maximum length of $512$.
The size of layout embedding is $768$.
The learning rates of fine-tuning are $1e^{-5}$ (chosen from $1e^{-3}$ to $1e^{-6}$).
We optimize our model with Adam and run it on one 32G Tesla V100 GPU for $30$ epochs, which takes 12/12 hours for the pre-training, and 1/0.5, 1/0.5, 1/0.5 hours for the fine-tuning of content extract, section extraction, relation extraction in product/official documents respectively.
All hyper-parameters are tuned on the dev set.

We intuitively name the layout-aware method \emph{LayoutBERT}. 
To access the effects of pre-training strategies, we implement two advanced methods, \emph{LayoutBERT (w/ MLLM)} and \emph{LayoutBERT (w/ MLLM, PHS)}.
Besides, we also report the results of \emph{LayoutLMv2}~\cite{xu2021layoutlmv2}, a representative state-of-the-art model on previous layout-aware IE datasets, as a strong baseline. It employs ResNeXt-FPN~\cite{xie2017aggregated} to generate image embeddings and concatenates them with text embeddings.
For fair comparison, \emph{LayoutLMv2} is also pre-trained in the same dataset with \emph{LayoutBERT}.

\noindent\textbf{Evaluation Metrics.}
Following popular choices, we adopt \emph{F1} score, the harmonic mean of \emph{precision} and \emph{recall}, for evaluation. Formally, precision and recall are defined as
$P = S_m / |\mathcal{P}|$, $R = S_m / |\mathcal{G}|$, where $S_m$ is the matching score of all predicted results, $\mathcal{|P|}$ and $\mathcal{|G|}$ are  the size of prediction set and ground truth set, respectively.
For \emph{hierarchy extraction} and \emph{relation extraction}, $S_m = \sum_{i}^{|\mathcal{P}|} m(p_i)$, where $m(p_i)$ is set to 1 if the $i$-th predicted result matches the ground truth strictly, otherwise it is 0.
For \emph{section extraction}, we calculate $S_m$, as the evaluation in Open IE systems, by computing the similarity between each predicted fact in $\mathcal{P}$ and each ground truth fact in $\mathcal{G}$, then find the optimal matching to maximize the sum of matched similarities by solving a linear assignment problem~\cite{sun2018logician}.
In the procedure, the similarity between two facts is defined as $s(p_i, g_j) = \sum_{l=1}^2\mathcal{M}(p_i^x, g_j^x)/2$, where $p_i^x$ and $g_j^x$ denote the $x$-th element (i.e., heading or body) of tuple $p_i$ and $g_j$, $\mathcal{M}(\cdot,\cdot)$ denotes the gestalt pattern matching measure~\cite{ratcliff1988pattern} for two strings.

\begin{table*}[t]\small
	\centering
	\caption{Performance of different models on LIE (\%). \emph{Product}, \emph{Official} and \emph{Average} refer to product documents, official documents and average results, respectively. The improvement over baseline is significant ($p$-value < 0.05).}
	\begin{tabular}{lccccccccc}
		\toprule
		\multirow{2}{*}{Models} & \multicolumn{3}{c}{Hierarchy Extraction} &  \multicolumn{3}{c}{Section Extraction} &  \multicolumn{3}{c}{Relation Extraction} \\
		     & \emph{Product} & \emph{Official} & \emph{Average} & \emph{Product} & \emph{Official} & \emph{Average} & \emph{Product} & \emph{Official} & \emph{Average} \\
		\midrule
		BERT~\cite{devlin2019bert} & 75.9 & 71.2 & 73.6 & 80.2 &  70.5 & 75.4 & 52.3 & 71.9 & 62.1 \\
		\midrule
		LayoutBERT               &  76.8 & 73.1 & 75.0 & 82.8 &  71.5 & 77.2 & 52.1 & 71.1  & 61.6 \\
		LayoutBERT (w/ MLLM)      & 77.8 & 75.9 & 76.9 & 84.2 & 72.8 & 78.5 & 53.9  & 75.8 & 64.9 \\
		LayoutBERT (w/ MLLM, PHS) & \textbf{78.3} & \textbf{76.5}  & \textbf{77.4} & \textbf{85.1} & \textbf{73.2}  & \textbf{79.2} & \textbf{54.2} & \textbf{76.7} & \textbf{65.5} \\
		\midrule
		LayoutLMv2~\cite{xu2021layoutlmv2} & 75.4 & 72.6 & 74.0 & 80.9 & 70.1 & 75.5 & 51.9 & 70.3 & 61.1 \\
		\bottomrule
	\end{tabular}
	\label{tab.main}
\end{table*}

\begin{figure*}[t]
\centering
\subfigure[Hierarchy extraction.]{
\includegraphics[width=0.25\linewidth]{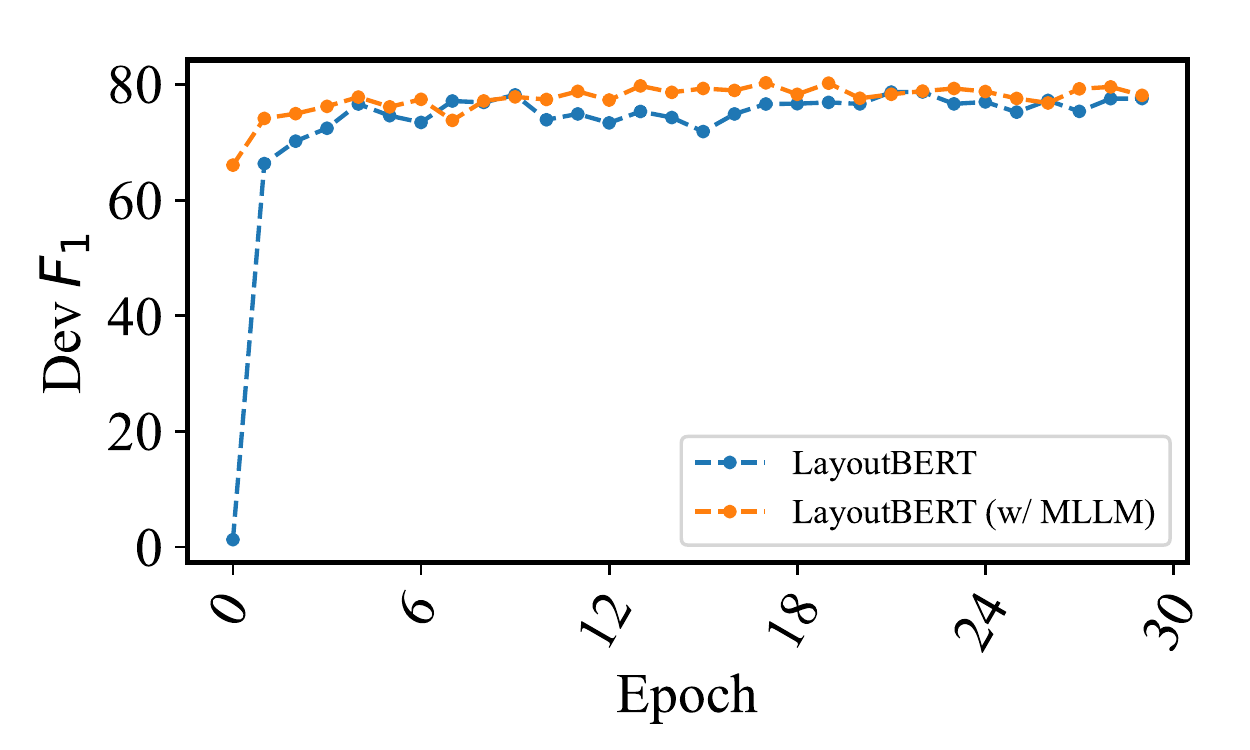}
\label{fig.task1.conv}
}
%\;
\subfigure[Section Extraction.]{
\includegraphics[width=0.25\linewidth]{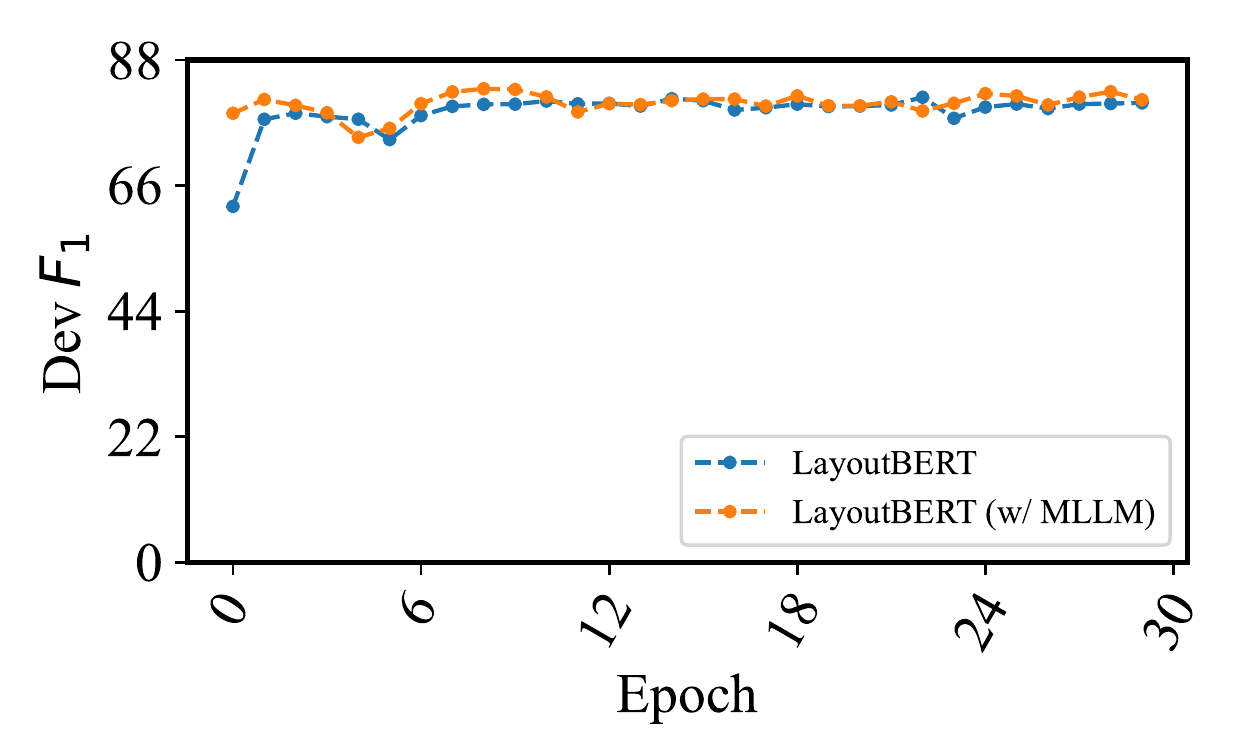}
\label{fig.task2.conv}
}
%\;
\subfigure[Relation Extraction.]{
\includegraphics[width=0.25\linewidth]{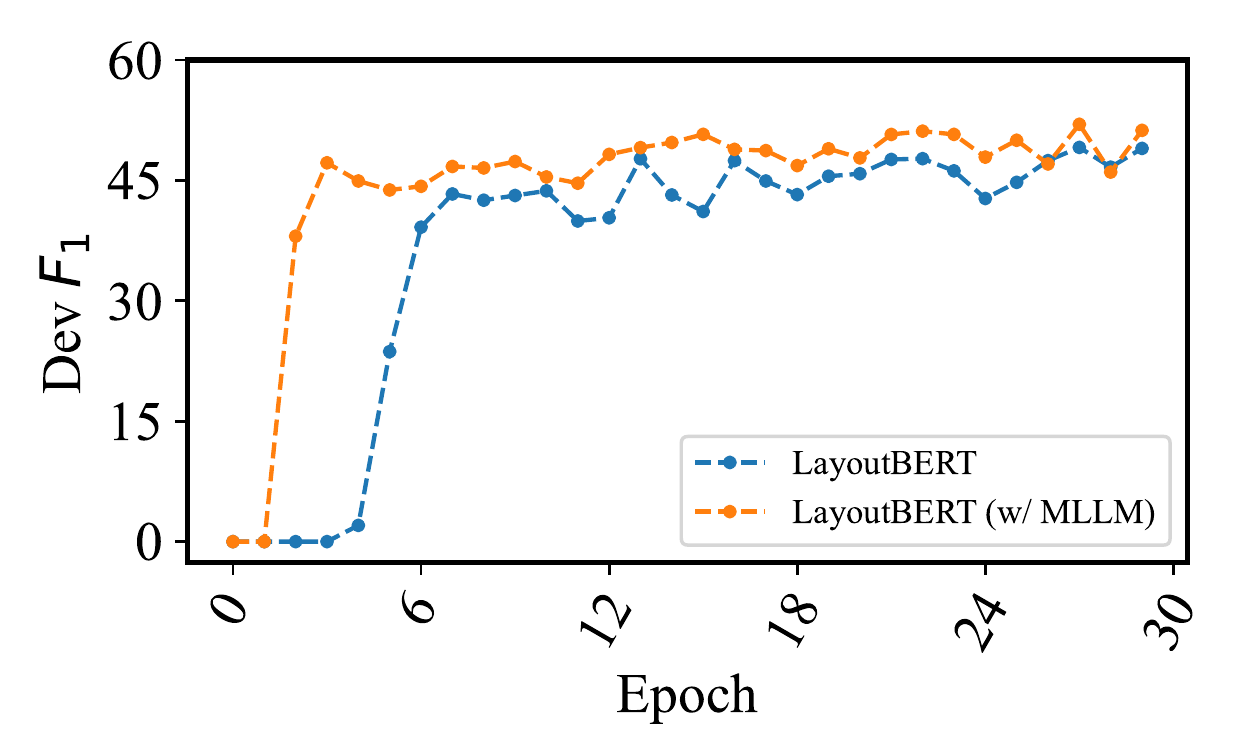}
\label{fig.task3.conv}
}
\caption{Convergence processes of layout-aware models on LIE (product documents) with and without in-domain pre-training. To minimize model variables, we compare LayoutBERT (w/ MLLM) and LayoutBERT here.}
\label{fig.pretrain}
\end{figure*}

\subsection{Results}

Comparing the performance of different models in Table~\ref{tab.main}, the first conclusion we draw is that layout-aware models outperform vanilla token-based models in almost all the tasks, which demonstrates the effectiveness of incorporating layout features with pre-trained language models.
Secondly, the process of in-domain pre-training further improves all of the download IE performance.
Through further analysis, we also find that pre-training dramatically accelerates the convergence in all downstream tasks (see Figure~\ref{fig.pretrain}).
Thirdly, LayoutLMv2 fails in this task and only achieves comparable results as BERT. Further study suggests that for such text-centered VRDs, the low-resolution image introduced by LayoutLMv2 becomes a noise interference rather than an information source.
Therefore, designing a more reasonable multi-modality integration method is also a point that needs to be studied in the future.

When looking at the results vertically, one can find that:
(1) Section extraction and relation extraction hold the highest and lowest performance, respectively.
(2) The performance gaps between these two kinds of documents should not be overlooked. 
We roughly attribute them to the difference of annotated data, including the number of tag sets and training samples.

\subsection{Analysis}

To understand the dataset and method more deeply, we carry out the model performance with respect to different aspects:

\begin{table}[t]\small
	\centering
	\caption{Performance w.r.t. input positional features on hierarchy extraction.}
	\begin{tabular}{lccccccccc}
		\toprule
		 & \emph{Product} & \emph{Official} & \emph{Average} \\
		\midrule
		LayoutBERT (w/ MLLM) & 77.8 & 75.9 & 76.9 \\
		\midrule
		\quad w/o 1D embedding & 49.5 & 57.4 & 53.5 \\
		\quad w/o 2D embedding & 77.0 & 73.6 & 75.3 \\
		\bottomrule
	\end{tabular}
	\label{tab.ablation}
\end{table}

\noindent\textbf{Performance w.r.t. Input Features.}
Table~\ref{tab.ablation} summarizes the ablation studies when removing different input features.
The results show that both 1D position embedding and 2D layout embedding contribute to the final performance, especially when we remove the 1D position embedding, the performance tends to collapse.
We believe that sequential information provides great help for document modeling, because there are many text fragments in the document and the 1D position reflects the reading order.
Besides, the model is hard to converge if text embedding is discarded.
It is contrary to the previous observation on form understanding dataset~\cite{cao2021extracting}, which again proves that the new dataset requires a powerful natural language understanding ability.

\begin{figure}[t]
\centering
\includegraphics[width=0.6\linewidth]{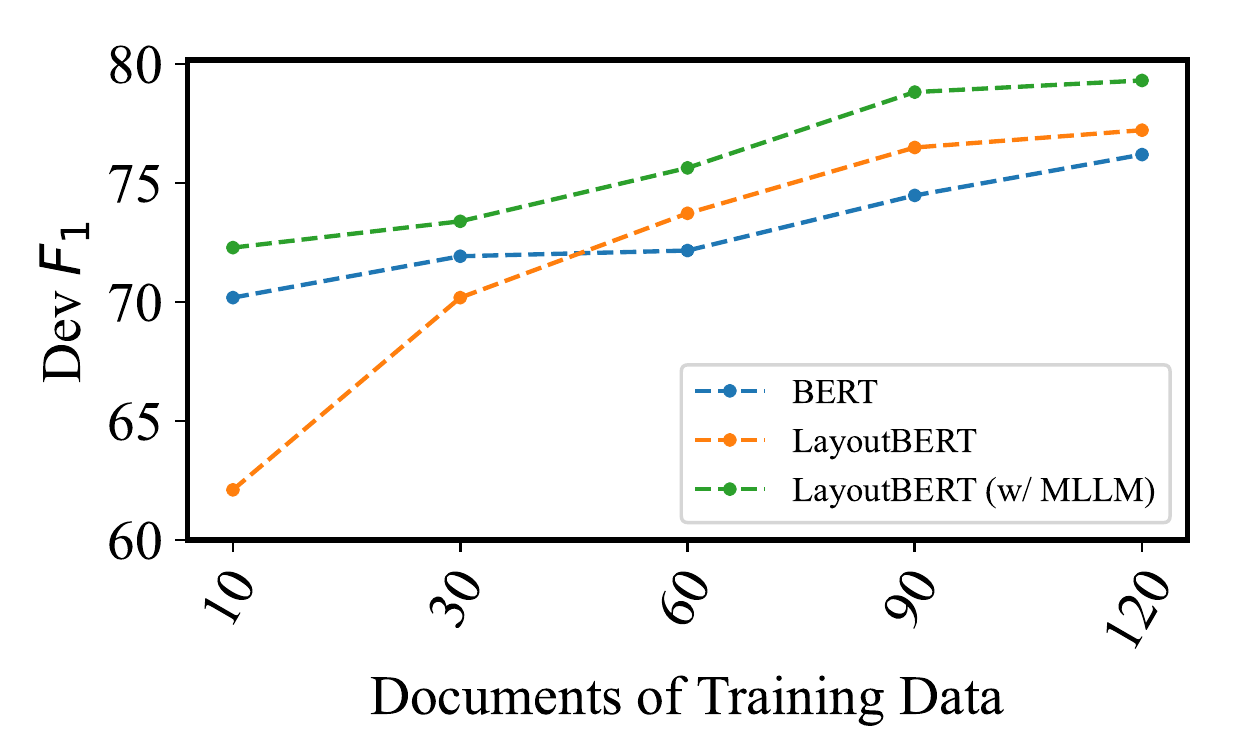}
\caption{Performance w.r.t. training data on hierarchy extraction (product documents).}
\label{fig.data}
\end{figure}

\noindent\textbf{Performance w.r.t. Training Data.}
We randomly sample \{10, 30, 60, 90, 120\} documents from the training set to explore the performance with different amounts of training data.
Figure~\ref{fig.data} shows that the performance gain of LayoutBERT relative to BERT increases first and then decreases. 
We analyze that the newly introduced parameters cannot be fully optimized with limited training data. When the data increases to enough, BERT is also capable of capturing some semantic information related to decision-making.
Unsurprisingly, the pre-training stage reduces the sensitivity of training data and brings stable improvements.

\begin{table}[t]\small
	\centering
	\caption{Performance w.r.t. domain transfer on section extraction. Row/column denote the train/test domains.}
	\begin{tabular}{lccccccccc}
		\toprule
		 & \emph{Product} & \emph{Official} & \emph{Average} \\
		\midrule
		\emph{Product}  & 85.1 & 9.8 & 47.5 \\
		\emph{Official}  & 30.5 & 73.2 & 51.9 \\
		\midrule
		\emph{All (Product+Official)} & 83.0 & 68.1 & 75.6 \\
		\bottomrule
	\end{tabular}
	\label{tab.transfer}
\end{table}

\begin{figure*}[t]
\centering
\includegraphics[width=\linewidth]{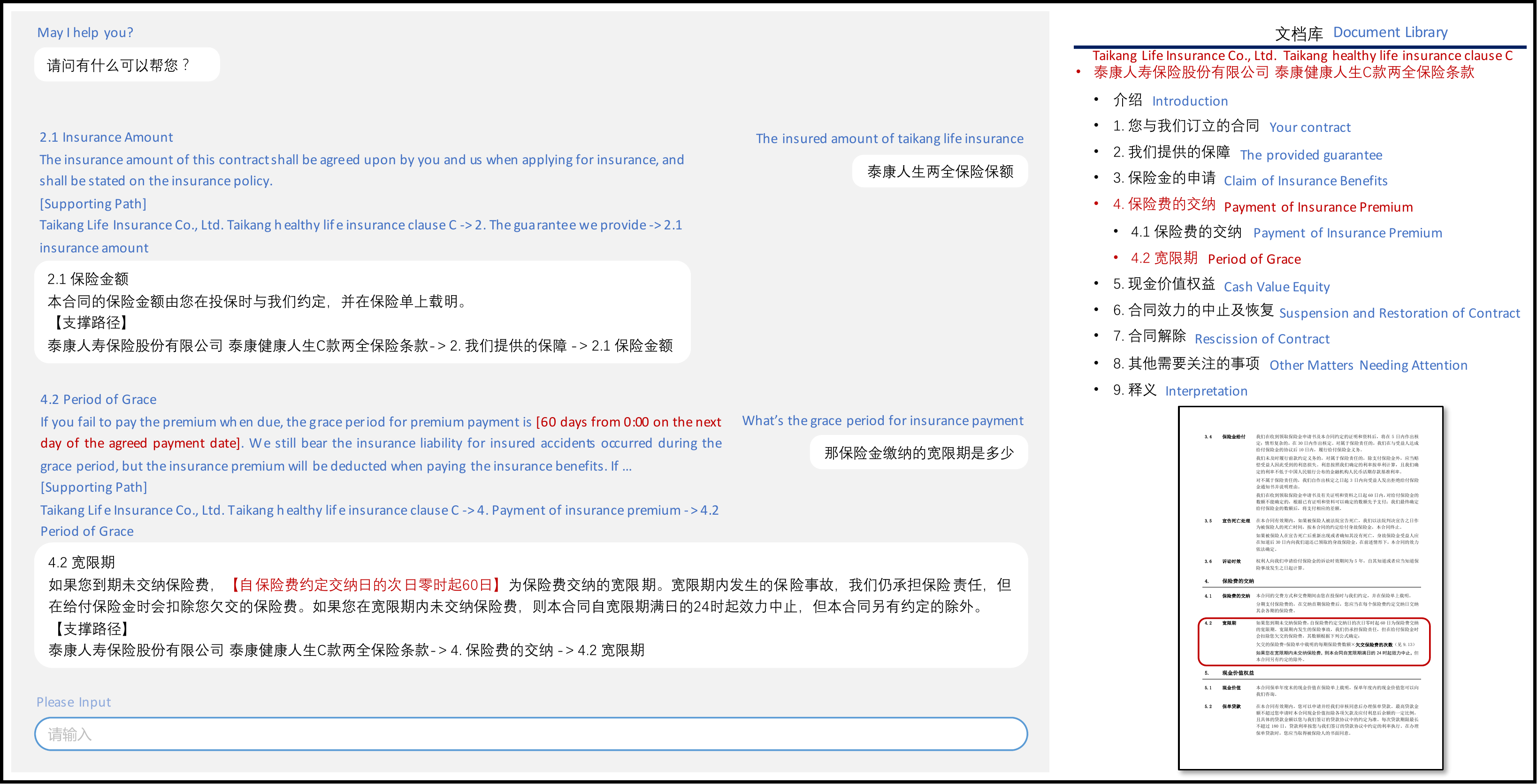}
\caption{Demonstration of the online doc2dial service. The left part is the interactive dialogue interface and the right shows current supporting path in relevant document, which is changes dynamically as user utterance change (The deployed system is in Chinese, and we add English annotations here for better understanding).}
\label{fig.demo}
\end{figure*}

\noindent\textbf{Performance w.r.t. Domain Transfer.}
For the exploration of domain transfer, we present evaluation results in Table~\ref{tab.transfer}. 
An interesting observation is that although the model is only trained on product documents, it can still transfer some knowledge to official documents, and vice versa.
Moreover, if documents from other domains are introduced in the training phase, it causes some negative impacts but reach a high average performance on the whole.
This verifies that LayoutBERT captures common layout and format invariance among different documents and transfer to other domains for document understanding.
Besides, introducing more data is a possible solution to further improve performance, even if new data comes from different fields or styles.

\subsection{Further Discussions}
We investigate the model outputs on dev set and find that there are some typical errors for current models:
(1) In \emph{hierarchy extraction}, the predicted results miss some section headings. For example, the model outputs the first and third section headings but misses the second heading, resulting in a continuous error after the second heading if we explicitly generate TOC number. It might be useful to introduce auxiliary training objectives to model headings with same format.
(2) In \emph{section extraction}, errors are mainly lie in the prediction of long section bodies (especially in official documents). Note that many bodies appear between two headings, so considering TOC of the document or replacing base encoder with a long-form language model (e.g., BigBird~\cite{zaheer2020big}) may be possible solutions.
(3) In \emph{relation extraction}, the model cannot work very well if two entities are far apart, which is the new challenge posed by LIE and accounts for a large proportion. In extreme cases, the head and tail entities are located at the document's beginning and end, respectively. 
We believe that introducing graph neural networks based on document structure and layout to explicitly enhance the correlation between entities will be a direction worthy of exploration.

\section{Demonstration}

Figure~\ref{fig.demo} shows the system demonstration of our doc2dial service in a real-world cloud computing platform, which is powered by the layout-aware IE model, a business document library, and a series of online dialogue-oriented algorithms~\cite{liu2021dialoguecse,dai2020learning}.
Following the doc2dial framework~\cite{feng2020doc2dialframe}, we perform knowledge extraction with the above three IE tasks in advance.
In online service, given an utterance like "\verb|The| \verb|insured| \verb|amount| \verb|of| \verb|taikang| \verb|life| \verb|insurance|", the doc2dial system first retrieves relevant document from \emph{Document Library}, and then returns a section-level coarse-grained response with supporting path, i.e., the location of this section in the document.
These two capabilities benefit from the off-line \emph{section extraction} and \emph{hierarchy extraction} processes, respectively.
In the next round, the user asked for "\verb|grace| \verb|period|", the system locates in section 4.2 by performing similar workflow, and highlights "\verb|60| \verb|days| \verb|from| \verb|0:00| \verb|on| \verb|the| \verb|next| \verb|day| \verb|of| \verb|the| \verb|agreed| \verb|payment| \verb|date|" as a fine-grained response, since the span was extracted from relation extraction in advance and it may be the information that the user is most concerned about.
To ensure the integrity and professionalism of response, we reserve all text of the answer section.

\section{Conclusion}

Nowadays, the digitalization and intellectualization of various industries are accelerating, we believe that mining and organizing valuable knowledge from massive data to enable downstream applications of enterprises is an important track for the industrial application of AI technology.
In this paper, we release a systematic novel layout-aware IE dataset and method for document-grounded dialogue systems, and highlight some possible future directions for VRD-based IE.
Experimental results suggest the effectiveness of incorporating pre-trained language models with layout features, and the demonstration confirms the importance of both structural and semantic knowledge in the doc2dial system.

\section*{Acknowledgments}
We would like to thank the anonymous reviewers for their insightful comments and constructive suggestions. 
This work is supported by the National Key Research and Development Program of China (grant No.2021YFB3100600), the Strategic Priority Research Program of Chinese Academy of Sciences (grant No.XDC02040400) and the Youth Innovation Promotion Association of Chinese Academy of Sciences (grant No.2021153).

\bibliographystyle{plainnat}
\bibliography{sample-base}

\begin{thebibliography}{42}
\providecommand{\natexlab}[1]{#1}
\providecommand{\url}[1]{\texttt{#1}}
\expandafter\ifx\csname urlstyle\endcsname\relax
  \providecommand{\doi}[1]{doi: #1}\else
  \providecommand{\doi}{doi: \begingroup \urlstyle{rm}\Url}\fi

\bibitem[Appalaraju et~al.(2021)Appalaraju, Jasani, Kota, Xie, and
  Manmatha]{appalaraju2021docformer}
Srikar Appalaraju, Bhavan Jasani, Bhargava~Urala Kota, Yusheng Xie, and
  R~Manmatha.
\newblock Docformer: End-to-end transformer for document understanding.
\newblock In \emph{ICCV}, 2021.

\bibitem[Campos et~al.(2020)Campos, Otegi, Soroa, Deriu, Cieliebak, and
  Agirre]{campos2020doqa}
Jon~Ander Campos, Arantxa Otegi, Aitor Soroa, Jan~Milan Deriu, Mark Cieliebak,
  and Eneko Agirre.
\newblock Doqa-accessing domain-specific faqs via conversational qa.
\newblock In \emph{ACL}, 2020.

\bibitem[Cao and Luo(2021)]{cao2021extracting}
Rongyu Cao and Ping Luo.
\newblock Extracting zero-shot structured information from form-like documents:
  Pretraining with keys and triggers.
\newblock In \emph{AAAI}, 2021.

\bibitem[Chen et~al.(2021)Chen, Lin, Zhou, Ma, Francis, Nyberg, and
  Oltramari]{chen2021building}
Xi~Chen, Faner Lin, Yeju Zhou, Kaixin Ma, Jonathan Francis, Eric Nyberg, and
  Alessandro Oltramari.
\newblock Building goal-oriented document-grounded dialogue systems.
\newblock In \emph{DialDoc}, 2021.

\bibitem[Chen et~al.(2020)Chen, Meng, Li, Chen, Xu, Xu, and
  Zhou]{chen2020bridging}
Xiuyi Chen, Fandong Meng, Peng Li, Feilong Chen, Shuang Xu, Bo~Xu, and Jie
  Zhou.
\newblock Bridging the gap between prior and posterior knowledge selection for
  knowledge-grounded dialogue generation.
\newblock In \emph{EMNLP}, pages 3426--3437, 2020.

\bibitem[Choi et~al.(2018)Choi, He, Iyyer, Yatskar, Yih, Choi, Liang, and
  Zettlemoyer]{choi2018quac}
Eunsol Choi, He~He, Mohit Iyyer, Mark Yatskar, Wen-tau Yih, Yejin Choi, Percy
  Liang, and Luke Zettlemoyer.
\newblock Quac: Question answering in context.
\newblock In \emph{EMNLP}, 2018.

\bibitem[Dai et~al.(2020)Dai, Li, Tang, Li, Sun, and Zhu]{dai2020learning}
Yinpei Dai, Hangyu Li, Chengguang Tang, Yongbin Li, Jian Sun, and Xiaodan Zhu.
\newblock Learning low-resource end-to-end goal-oriented dialog for fast and
  reliable system deployment.
\newblock In \emph{ACL}, pages 609--618, 2020.

\bibitem[Devlin et~al.(2019)Devlin, Chang, Lee, and Toutanova]{devlin2019bert}
Jacob Devlin, Ming-Wei Chang, Kenton Lee, and Kristina Toutanova.
\newblock Bert: Pre-training of deep bidirectional transformers for language
  understanding.
\newblock In \emph{NAACL}, 2019.

\bibitem[Fadnis et~al.(2021)Fadnis, Dhoolia, Zhu, Liao, Ross, Mills, Joshi, and
  Lastras]{fadnis2021doc2bot}
Kshitij Fadnis, Pankaj Dhoolia, Li~Zhu, Q~Vera Liao, Steven Ross, Nathaniel
  Mills, Sachindra Joshi, and Luis Lastras.
\newblock Doc2bot: Document grounded bot framework.
\newblock In \emph{AAAI}, 2021.

\bibitem[Feng et~al.(2020{\natexlab{a}})Feng, Fadnis, Liao, and
  Lastras]{feng2020doc2dialframe}
Song Feng, Kshitij Fadnis, Q~Vera Liao, and Luis~A Lastras.
\newblock Doc2dial: a framework for dialogue composition grounded in documents.
\newblock In \emph{AAAI}, 2020{\natexlab{a}}.

\bibitem[Feng et~al.(2020{\natexlab{b}})Feng, Wan, Gunasekara, Patel, Joshi,
  and Lastras]{feng2020doc2dialdata}
Song Feng, Hui Wan, Chulaka Gunasekara, Siva Patel, Sachindra Joshi, and Luis
  Lastras.
\newblock doc2dial: A goal-oriented document-grounded dialogue dataset.
\newblock In \emph{EMNLP}, 2020{\natexlab{b}}.

\bibitem[He et~al.(2022{\natexlab{a}})He, Dai, Yang, Sun, Huang, Si, and
  Li]{he2022sapce}
Wanwei He, Yinpei Dai, Min Yang, Jian Sun, Fei Huang, Luo Si, and Yongbin Li.
\newblock Unified dialog model pre-training for task-oriented dialog
  understanding and generation.
\newblock In \emph{SIGIR}, page 187–200. Association for Computing Machinery,
  2022{\natexlab{a}}.
\newblock ISBN 9781450387323.

\bibitem[He et~al.(2022{\natexlab{b}})He, Dai, Zheng, Wu, Cao, Liu, Jiang,
  Yang, Huang, Si, et~al.]{he2022galaxy}
Wanwei He, Yinpei Dai, Yinhe Zheng, Yuchuan Wu, Zheng Cao, Dermot Liu, Peng
  Jiang, Min Yang, Fei Huang, Luo Si, et~al.
\newblock Galaxy: A generative pre-trained model for task-oriented dialog with
  semi-supervised learning and explicit policy injection.
\newblock In \emph{AAAI}, volume~36, pages 10749--10757, 2022{\natexlab{b}}.

\bibitem[Hendrickx et~al.(2010)Hendrickx, Kim, Kozareva, Nakov, S{\'e}aghdha,
  Pad{\'o}, Pennacchiotti, Romano, and Szpakowicz]{hendrickx2010semeval}
Iris Hendrickx, Su~Nam Kim, Zornitsa Kozareva, Preslav Nakov, Diarmuid~{\'O}
  S{\'e}aghdha, Sebastian Pad{\'o}, Marco Pennacchiotti, Lorenza Romano, and
  Stan Szpakowicz.
\newblock Semeval-2010 task 8: Multi-way classification of semantic relations
  between pairs of nominals.
\newblock In \emph{SemEval}, 2010.

\bibitem[Huang et~al.(2019)Huang, Chen, He, Bai, Karatzas, Lu, and
  Jawahar]{huang2019icdar2019}
Zheng Huang, Kai Chen, Jianhua He, Xiang Bai, Dimosthenis Karatzas, Shijian Lu,
  and CV~Jawahar.
\newblock Icdar2019 competition on scanned receipt ocr and information
  extraction.
\newblock In \emph{ICDAR}, 2019.

\bibitem[Jain et~al.(2020)Jain, van Zuylen, Hajishirzi, and
  Beltagy]{jain2020scirex}
Sarthak Jain, Madeleine van Zuylen, Hannaneh Hajishirzi, and Iz~Beltagy.
\newblock Scirex: A challenge dataset for document-level information
  extraction.
\newblock In \emph{ACL}, 2020.

\bibitem[Jaume et~al.(2019)Jaume, Ekenel, and Thiran]{jaume2019funsd}
Guillaume Jaume, Hazim~Kemal Ekenel, and Jean-Philippe Thiran.
\newblock Funsd: A dataset for form understanding in noisy scanned documents.
\newblock In \emph{ICDAR: Workshops}, 2019.

\bibitem[Khosla et~al.(2021)Khosla, Lovelace, Dutt, and
  Pratapa]{khosla2021team}
Sopan Khosla, Justin Lovelace, Ritam Dutt, and Adithya Pratapa.
\newblock Team jars: Dialdoc subtask 1-improved knowledge identification with
  supervised out-of-domain pretraining.
\newblock In \emph{DialDoc}, 2021.

\bibitem[Li et~al.(2021{\natexlab{a}})Li, Bi, Yan, Wang, Huang, Huang, and
  Si]{li2021structurallm}
Chenliang Li, Bin Bi, Ming Yan, Wei Wang, Songfang Huang, Fei Huang, and Luo
  Si.
\newblock Structurallm: Structural pre-training for form understanding.
\newblock In \emph{ACL}, 2021{\natexlab{a}}.

\bibitem[Li et~al.(2021{\natexlab{b}})Li, Qian, Yu, Qin, Zhang, Liu, Yao, Han,
  Liu, and Ding]{li2021structext}
Yulin Li, Yuxi Qian, Yuechen Yu, Xiameng Qin, Chengquan Zhang, Yan Liu, Kun
  Yao, Junyu Han, Jingtuo Liu, and Errui Ding.
\newblock Structext: Structured text understanding with multi-modal
  transformers.
\newblock In \emph{ACM MM}, 2021{\natexlab{b}}.

\bibitem[Liu et~al.(2021)Liu, Wang, Liu, Sun, Huang, and
  Si]{liu2021dialoguecse}
Che Liu, Rui Wang, Jinghua Liu, Jian Sun, Fei Huang, and Luo Si.
\newblock Dialoguecse: Dialogue-based contrastive learning of sentence
  embeddings.
\newblock In \emph{EMNLP}, pages 2396--2406, 2021.

\bibitem[Park et~al.(2019)Park, Shin, Lee, Lee, Surh, Seo, and
  Lee]{park2019cord}
Seunghyun Park, Seung Shin, Bado Lee, Junyeop Lee, Jaeheung Surh, Minjoon Seo,
  and Hwalsuk Lee.
\newblock Cord: A consolidated receipt dataset for post-ocr parsing.
\newblock In \emph{NeurIPS: Workshop}, 2019.

\bibitem[Ratcliff and Metzener(1988)]{ratcliff1988pattern}
John~W Ratcliff and David~E Metzener.
\newblock Pattern matching: The gestalt approach.
\newblock \emph{Dr Dobbs Journal}, 13\penalty0 (7):\penalty0 46, 1988.

\bibitem[Sun et~al.(2018)Sun, Li, Wang, Fan, Feng, and Li]{sun2018logician}
Mingming Sun, Xu~Li, Xin Wang, Miao Fan, Yue Feng, and Ping Li.
\newblock Logician: A unified end-to-end neural approach for open-domain
  information extraction.
\newblock In \emph{WSDM}, 2018.

\bibitem[Wan et~al.(2021)Wan, Feng, Gunasekara, Patel, Joshi, and
  Lastras]{wan2021does}
Hui Wan, Song Feng, Chulaka Gunasekara, Siva~Sankalp Patel, Sachindra Joshi,
  and Luis Lastras.
\newblock Does structure matter? encoding documents for machine reading
  comprehension.
\newblock In \emph{NAACL}, 2021.

\bibitem[Wang et~al.(2021)Wang, Liu, Jin, Tang, Zhang, Zhang, Wang, Wu, and
  Cai]{wang2021towards}
Jiapeng Wang, Chongyu Liu, Lianwen Jin, Guozhi Tang, Jiaxin Zhang, Shuaitao
  Zhang, Qianying Wang, Yaqiang Wu, and Mingxiang Cai.
\newblock Towards robust visual information extraction in real world: New
  dataset and novel solution.
\newblock In \emph{AAAI}, 2021.

\bibitem[Wang et~al.(2020)Wang, Yu, Zhang, Liu, Zhu, and Sun]{wang2020tplinker}
Yucheng Wang, Bowen Yu, Yueyang Zhang, Tingwen Liu, Hongsong Zhu, and Limin
  Sun.
\newblock Tplinker: Single-stage joint extraction of entities and relations
  through token pair linking.
\newblock In \emph{COLING}, pages 1572--1582, 2020.

\bibitem[Wei et~al.(2020)Wei, He, and Zhang]{wei2020robust}
Mengxi Wei, Yifan He, and Qiong Zhang.
\newblock Robust layout-aware ie for visually rich documents with pre-trained
  language models.
\newblock In \emph{SIGIR}, 2020.

\bibitem[Wei et~al.(2018)Wei, Liu, Peng, Tou, Chen, Huang, Wong, and
  Dai]{wei2018task}
Zhongyu Wei, Qianlong Liu, Baolin Peng, Huaixiao Tou, Ting Chen, Xuan-Jing
  Huang, Kam-Fai Wong, and Xiang Dai.
\newblock Task-oriented dialogue system for automatic diagnosis.
\newblock In \emph{ACL}, 2018.

\bibitem[Wolf et~al.(2020)Wolf, Chaumond, Debut, Sanh, Delangue, Moi, Cistac,
  Funtowicz, Davison, Shleifer, et~al.]{wolf2020transformers}
Thomas Wolf, Julien Chaumond, Lysandre Debut, Victor Sanh, Clement Delangue,
  Anthony Moi, Pierric Cistac, Morgan Funtowicz, Joe Davison, Sam Shleifer,
  et~al.
\newblock Transformers: State-of-the-art natural language processing.
\newblock In \emph{EMNLP: Demo}, 2020.

\bibitem[Xie et~al.(2017)Xie, Girshick, Doll{\'a}r, Tu, and
  He]{xie2017aggregated}
Saining Xie, Ross Girshick, Piotr Doll{\'a}r, Zhuowen Tu, and Kaiming He.
\newblock Aggregated residual transformations for deep neural networks.
\newblock In \emph{CVPR}, 2017.

\bibitem[Xu et~al.(2021)Xu, Xu, Lv, Cui, Wei, Wang, Lu, Florencio, Zhang, Che,
  et~al.]{xu2021layoutlmv2}
Yang Xu, Yiheng Xu, Tengchao Lv, Lei Cui, Furu Wei, Guoxin Wang, Yijuan Lu,
  Dinei Florencio, Cha Zhang, Wanxiang Che, et~al.
\newblock Layoutlmv2: Multi-modal pre-training for visually-rich document
  understanding.
\newblock In \emph{ACL}, 2021.

\bibitem[Xu et~al.(2020)Xu, Li, Cui, Huang, Wei, and Zhou]{xu2020layoutlm}
Yiheng Xu, Minghao Li, Lei Cui, Shaohan Huang, Furu Wei, and Ming Zhou.
\newblock Layoutlm: Pre-training of text and layout for document image
  understanding.
\newblock In \emph{KDD}, 2020.

\bibitem[Yao et~al.(2019)Yao, Ye, Li, Han, Lin, Liu, Liu, Huang, Zhou, and
  Sun]{yao2019docred}
Yuan Yao, Deming Ye, Peng Li, Xu~Han, Yankai Lin, Zhenghao Liu, Zhiyuan Liu,
  Lixin Huang, Jie Zhou, and Maosong Sun.
\newblock Docred: A large-scale document-level relation extraction dataset.
\newblock In \emph{ACL}, 2019.

\bibitem[Yu et~al.(2020{\natexlab{a}})Yu, Zhang, Shu, Liu, Wang, Wang, and
  Li]{yu2020joint}
Bowen Yu, Zhenyu Zhang, Xiaobo Shu, Tingwen Liu, Yubin Wang, Bin Wang, and
  Sujian Li.
\newblock Joint extraction of entities and relations based on a novel
  decomposition strategy.
\newblock In \emph{ECAI}, pages 2282--2289. IOS Press, 2020{\natexlab{a}}.

\bibitem[Yu et~al.(2020{\natexlab{b}})Yu, Sun, Cardie, and Yu]{yu2020dialogue}
Dian Yu, Kai Sun, Claire Cardie, and Dong Yu.
\newblock Dialogue-based relation extraction.
\newblock In \emph{ACL}, 2020{\natexlab{b}}.

\bibitem[Zaheer et~al.(2020)Zaheer, Guruganesh, Dubey, Ainslie, Alberti,
  Ontanon, Pham, Ravula, Wang, Yang, and Ahmed]{zaheer2020big}
Manzil Zaheer, Guru Guruganesh, Kumar~Avinava Dubey, Joshua Ainslie, Chris
  Alberti, Santiago Ontanon, Philip Pham, Anirudh Ravula, Qifan Wang, Li~Yang,
  and Amr Ahmed.
\newblock Big bird: Transformers for longer sequences.
\newblock In \emph{NeurIPS}, 2020.

\bibitem[Zhang et~al.(2020{\natexlab{a}})Zhang, Xu, Cheng, Pu, Lu, Qiao, Niu,
  and Wu]{zhang2020trie}
Peng Zhang, Yunlu Xu, Zhanzhan Cheng, Shiliang Pu, Jing Lu, Liang Qiao, Yi~Niu,
  and Fei Wu.
\newblock Trie: End-to-end text reading and information extraction for document
  understanding.
\newblock In \emph{ACM MM}, 2020{\natexlab{a}}.

\bibitem[Zhang et~al.(2020{\natexlab{b}})Zhang, Yu, Shu, Liu, Tang, Yubin, and
  Guo]{zhang2020document}
Zhenyu Zhang, Bowen Yu, Xiaobo Shu, Tingwen Liu, Hengzhu Tang, Wang Yubin, and
  Li~Guo.
\newblock Document-level relation extraction with dual-tier heterogeneous
  graph.
\newblock In \emph{COLING}, pages 1630--1641, 2020{\natexlab{b}}.

\bibitem[Zhang et~al.(2021)Zhang, Yu, Shu, and Liu]{zhang2021aware}
Zhenyu Zhang, Bowen Yu, Xiaobo Shu, and Tingwen Liu.
\newblock Na-aware machine reading comprehension for document-level relation
  extraction.
\newblock In \emph{ECML/PKDD}, pages 580--595, 2021.

\bibitem[Zheng et~al.(2017)Zheng, Wang, Bao, Hao, Zhou, and Xu]{zheng2017joint}
Suncong Zheng, Feng Wang, Hongyun Bao, Yuexing Hao, Peng Zhou, and Bo~Xu.
\newblock Joint extraction of entities and relations based on a novel tagging
  scheme.
\newblock In \emph{ACL}, 2017.

\bibitem[Zhou et~al.(2022)Zhou, Yu, Sun, Long, Li, Sun, and
  Yongbin]{zhou2022survey}
Shaowen Zhou, Bowen Yu, Aixin Sun, Cheng Long, Jingyang Li, Jian Sun, and
  Li~Yongbin.
\newblock A survey on neural open information extraction: Current status and
  future directions.
\newblock In \emph{IJCAI}, 2022.

\end{thebibliography}

\appendix

\end{document}